\documentclass[twoside]{article}

\usepackage[accepted]{aistats2023}
\usepackage{microtype}
\usepackage{graphicx}
\usepackage{subcaption}
\usepackage{booktabs} 
\usepackage[linesnumbered,ruled,vlined]{algorithm2e}

\usepackage{amssymb}
\usepackage{amsthm}
\usepackage{graphicx}

\usepackage[round]{natbib}

\numberwithin{equation}{section}
\theoremstyle{plain}

\theoremstyle{definition}

\theoremstyle{remark}

\begin{document}

%

%

\twocolumn[

\aistatstitle{Nonparametric Probabilistic Regression with Coarse Learners}

\aistatsauthor{Brian Lucena}

\aistatsaddress{Numeristical} ]

\begin{abstract}
Probabilistic Regression refers to predicting a full probability density function for the target conditional on the features.  We present a nonparametric approach to this problem which combines base classifiers (typically gradient boosted forests) trained on different coarsenings of the target value.  By combining such classifiers and averaging the resulting densities, we are able to compute precise conditional densities with minimal assumptions on the shape or form of the density. We combine this approach with a structured cross-entropy loss function which serves to regularize and smooth the resulting densities.   Prediction intervals computed from these densities are shown to have high fidelity in practice.  Furthermore, examining the properties of these densities on particular observations can provide valuable insight.  We demonstrate this approach on a variety of datasets and show competitive performance, particularly on larger datasets. 
 \end{abstract}

\section{INTRODUCTION}
In modern machine learning parlance, the term {\em regression} conveys the prediction of a point estimate of numerical target.  While classical linear regression models can be thought of as predicting a distribution (based on Gaussian noise), the underlying assumptions behind this density are rarely true in practice.  As such, linear regression is typically used for point estimates.  As modern machine learning approaches to regression have gained in popularity, they have still primarily focused on point prediction as measured by root mean-squared error (RMSE).  The problem of {\em probabilistic regression} attempts to specify a full conditional probability density for the target variable given the features which is measurably realistic.  This is a natural next step in sophistication, enabled by larger datasets and more complex machine learning methods.

There are many reasons why knowing a precise conditional density is superior to a point estimate.  For example, in predicting an individual's rating of a restaurant, there is a big difference between a bimodal  and unimodel distributions with the same mean of 5 out of 10.  The former indicates a polarity of opinion (``you'll love it or you'll hate it") versus the assured mediocrity of the latter.  In a more serious matter, consider a cancer patient choosing between two treatments both of which, on average, add 2 years of longevity.  Certainly, if one of them is ineffective 80\% of the time but adds 10 years when effective, this is a different personal calculation than one which almost certainly extends life about two years.  This would be even more important if a treatment had significant probability of decreasing lifespan.

Most approaches to probabilistic regression either assume a particular parametric form, and/or require complicated and expensive computation to implement, rendering them inaccessible to all but experts in the particular method.  This paper presents a nonparametric method which makes minimal assumptions about the shape or form of the conditional probability density function.  Yet, to a practitioner, it is nearly as simple as fitting a standard gradient boosting or random forest model.  It is sophisticated enough to capture predicted densities with multiple modes and nonstandard density shapes, which can prove enlightening when visualized on real problems.  Moreover, having a full conditional density permits the user to define prediction intervals, which we empirically show to provide appropriate coverage in real-world examples.

In between the extremes of point prediction and probabilistic regression, there are several different approaches to give some indication of the level of uncertainty around the prediction.  Examples of this ``middle ground" include quantile regression, conformal prediction, and other methods to output confidence intervals around a prediction.

\section{RELATED WORK}
There are multiple techniques for probabilistic regression in the strictest sense - that is, giving a precise probability density function for the target value given the features.  NGBoost (\cite{ngboost}) is the most direct analog to the method presented here, as it is simple to implement, usable by non-experts, and also relies on gradient boosting.  However, our method is nonparametric, while NGBoost requires a parametric specification and then predicts the value of the parameters given the features.  Generalized Additive Models for Shape, Scale, and Location (GAMLSS) (\cite{gamlss}) are similarly restricted to a specified model form.   Bayesian approaches such as Bayesian Additive Regression Trees (BART) (~\cite{chipman2010bart}) require computationally expensive methods to sample the posterior distribution and are generally difficult to use.  Bayesian Deep Learning methods (e.g.~\cite{pmlr-v37-blundell15},~\cite{pmlr-v37-hernandez-lobatoc15},~\cite{graves2011}) have shown some promise but are also relatively difficult to use.

The approach of conformal prediction (\cite{ShaferVovk}) is related to probabilistic regression in that both attempt to go beyond a mere point estimate on regression problems.  Conformal prediction uses any machine learning model in conjunction with a user-specified confidence level to predict a range of values for the target, such that the target falls into the interval with the appropriate confidence.  However, it differs from probabilistic regression in that it does not explicitly output a probability density function.  Similarly, approaches like Quantile Regression (\cite{qr-koenker}), in conjunction with Bayesian Methods (\cite{bayesqr1},~\cite{bayesqr2}), Random Forests (\cite{meinshausen2006quantile}), or Neural Networks (\cite{qcnn},\cite{rodrigues2020beyond}) have been attempted, but do not output a full conditional probability density function and/or are not straightforward to use.

\section{CONTRIBUTIONS}

We present a nonparametric approach to probabilistic regression called PRESTO (Probabilistic REgression with Structured Trees).  Like NGBoost, it relies on gradient boosting (\cite{Friedman00greedyfunction}, \cite{chen2016xgboost}) and is easy to implement and fit on new data.  Specifically, models can be fit with the same ease as standard gradient boosting or random forest models using the standard fit/predict paradigm.  However, unlike NGBoost, PRESTO does not make any assumptions about the form of the conditional density.  Rather, it approximates the true conditional density by fitting multiple piecewise constant densities and averaging the results.  Consequently, it performs well in situations where the true conditional density is multimodal or otherwise does not have a simple or consistent parametric form.

We run the PRESTO algorithm on a benchmark collection of 10 datasets from the UC-Irvine repository and show that its performance compares favorably to the competitors.  Since we have a full conditional density for each prediction, we can directly create confidence intervals for the predicted values.  We show that, in practice, these intervals provide at least the guaranteed level of coverage.  Analysis of the predicted densities can provide useful insight into scientific problems, as we are able to detect multi-modality and other interesting effects in the conditional distribution.  Furthermore, we show that using a structured cross-entropy loss for the underlying gradient boosting base classifiers improves performance both quantitatively and qualitatively.   Finally, we examine learning curves to show that relatively few forests are needed to achieve near-optimal performance.

\section{PRESTO ALGORITHM}
Put simply, the PRESTO Algorithm builds multiple classifiers, converts their predictions to densities, and averages the results.  For each classifier, we randomly subdivide the range of the target variable into intervals, thus converting the regression problem into one of multi-classfication.  The resulting classifier yields a probability distribution over the various intervals which can be cast as a probability density function (pdf) that is piecewise constant.  By building many probabilistic classifiers on different choices of intervals, we can average the resulting distributions yielding a final probability density function.  The resulting pdf is still piecewise constant, but with such fine grain intervals that it typically resembles a curve, yet also handles discontinuities in the density function if they are present.

While the approach is straightforward, there are many details that must be handled elegantly for it to work in practice.  In particular, how are the random intervals chosen?  For some applications, there may be an a priori known range of possible values.  However, in general we must learn appropriate choices of intervals from the training data itself.  Moreover, even within a fixed range, there may be more resolution required in some areas than others, meaning that a fixed-width grid is not sufficient.  Further complications arise when you consider that there may be target values on future (test) observations that exceed the range seen in the test set.  Thus we may want to extend the range beyond what is seen in the training data so as to avoid assigning those extreme values a zero probability.

\subsection{Formal Description of PRESTO Algorithm}

\begin{algorithm}[t]
\label{alg:presto}
\DontPrintSemicolon
  \SetKwInOut{Input}{Input}\SetKwInOut{Output}{Output}
  \KwData{Training dataset $\{x_i, y_i\}_{i=1}^n$}
 \Input{number of classifiers: $m$: Base classifier: $\mathcal{C}$, interval selection method: {\em INTMETHOD}}
  \Output{A probabilistic regressor $f$}
  \For{$i \leftarrow 1, \ldots, m$}
  {Use INTMETHOD to select a sequence of interval endpoints: $b_0, b_1, \ldots, b_k$ with $b_j < b_{j+1}$.\;
  Create a discretized version $\{y^{\prime}_i\}$ of $\{y_i\}$ where $y^\prime = j$ iff $b_j \leq y_i <b_{j+1}$ (and $y^{\prime}_i = k$ if $y_i = b_k)$\;
  $g_i\leftarrow \mathcal{C}(\{x_i, y^\prime_i\}_{i=1}^n)$ \;
  $f_i(x) \leftarrow \mathcal{D}(g_i(x), B)$\;
  }

$f = avg(f_1, f_2, \ldots, f_m) $
\caption{PRESTO}
\end{algorithm}

We assume a training dataset of size $n$, $\{x_i, y_i\}_{i=1}^n$ where $x_i \in \mathcal{X}$ (the feature space) and $y_i \in \mathbb{R}$.  Let $\mathcal{P}$ represent the space of probability density functions (pdfs) on $\mathbb{R}$. We define a {\em probabilistic regressor} as a mapping $f: \mathcal{X} \rightarrow \mathcal{P}$.  Given a collection of $m$  probabilistic regressors, we can {\em average} them by defining $f(x) = avg(f_1, f_2, \ldots, f_m) := \frac{1}{m} \sum_{i=1}^m f_i(x)$ where the pdfs are summed in the obvious (pointwise) manner. Let $g$ be a (probabilistic) multi-classifier where the classes are represented by $\{1,2,\ldots,k\}$.  In this way, $g$ can be thought of as a mapping $g: \mathcal{X} \rightarrow \mathcal{S}_k$ where $S_k$ is the $(k-1)$-dimensional probability simplex (on $k$ classes).  A classifier method $\mathcal{C}$ when {\em fit} on a training dataset $\{x_i, y_i^{\prime}\}$ with $y_i^{\prime} \in \{1,2,\ldots,k\}$ yields a probabilistic multi-classifier, denoted $\mathcal{C}(\{x_i, y_i^{\prime}\})$.

Given a probability distribution $P =(p_1, p_2, \ldots, p_k)$ on $k$ classes, and a set of $k$ intervals $B = ([b_0, b_1],[b_1, b_2],\ldots,[b_{k-1},b_k])$ with $b_i < b_{i+1}$, define the associated density $\mathcal{D}(P, B)$ as:
\[ \mathcal{D}(P, B)= 
\begin{cases}
     p_i/(b_i - b_{i-1}),& \text{if } x \in [b_{i-1}, b_i) \\
     p_k/(b_k - b_{k-1}), & \text{if } x = b_k \\
         0,              & \text{otherwise}
\end{cases} \]
Clearly if $\sum_{i=1}^k p_i = 1$ then $\int_{-\infty}^{\infty} \mathcal{D}(P,B) = 1$, so $\mathcal{D}(P,B)$ is a valid probability density function.

The PRESTO algorithm is defined formally as Algorithm 1.  We must specify a base classifier, a number of classifiers $m$ and the interval selection method (INTMETHOD).  While any base classifier could be used in principle, in this paper we exclusively use forests of Gradient Boosted Decision Trees.  The interval selection method requires more nuance.  As discussed earlier, the distribution of the numerical target, and what is known about it a priori, may influence how we choose intervals.  Generally speaking, if the target is known a priori to take values only in a fixed interval and a consistent resolution is appropriate across that interval, we can just use the same specified set of intervals each time.  We refer to this as the {\em fixed} interval selection method.  For example, in the {\em Wine} dataset, the target is known to be integers between 2 and 8, inclusive.  Therefore, we can use a fixed set of points $\{2, 2.5, 3.5, \ldots,7.5,8\}$ to define our intervals.  We refer to this as the {\em fixed} method.  Alternatively, there may be a fixed range of values which is larger in cardinality.  For example in the {\em Naval} dataset, the target is known a priori to take values in $\{.95, .951, .952, \ldots, .999, 1\}$.  We could use a fixed set of points such as $\{.950, .9505, .9515, .9525, \ldots, .9985, .9995, 1\}$ to define intervals.  However, using the base classifier on 51 classes may not be most effective.  So we may want to choose only a (random) subset of those grid points (keeping the min and the max) to define the intervals.  We refer to this as the {\em fixed-rss} method.

In the most general case, we know nothing about the possible range of target values beyond what we see in the training data.  In these situations, we use the {\em RandQuantile} interval selection method, given as Algorithm 2.  In this method, we choose a subset of quantile values from the uniform distribution and then find the corresponding quantiles in the empirical distribution of the target training values.  After sorting these values (removing duplicates) and adding in the minimum and maximum observed values, we take the {\em midpoints} of the neighboring values (keeping the min and max).  This is done so that the observed y-values fall in the center of the bin rather than at an endpoint.

\begin{algorithm}[t]
\DontPrintSemicolon
  \SetKwInOut{Input}{Input}\SetKwInOut{Output}{Output}
  \KwData{Training dataset $\{x_i, y_i\}_{i=1}^n$}
  \Input{Number of quantiles to sample: $r$, Boolean: {\em Extend}, Extend params: $(q_{\min},q_{max},t)$}
 \Output{A set of interval endpoints: $\{b_0, b_1, \ldots, b_k\}$ with $b_j < b_{j+1}$}
Let $z_1, \ldots, z_r \sim U[0,1]$ {\em iid} and sorted s.t. $z_j <z_{j+1}$ \;
Let $a_j$ be the $z_j$-quantile of $\{y_i\}$. (for $j$ in $1, \ldots, r$) \;
Let $a_0, a_{r+1}$ be the min (resp. max) value in $\{y_i\}$ \;
Let $A = \{ a_0, (a_0+a_1)/2,(a_1+a_2)/2, \ldots, (a_r+a_{r+1})/2, a_{r+1}\}$ \;
\If{Extend}{$u \rightarrow$ $q_{\min}$-th quantile of $\{y_i\}$\;
		$v \rightarrow$ $q_{\max}$-th quantile of $\{y_i\}$\;
		$w \rightarrow (v-u)*t$\;
		$A \rightarrow A \cup \{y_{\min} - w, y_{\max} + w\} $}
Let $b_0, b_1, \ldots, b_k$ be the sorted, unique values of $A$  \;
\caption{RandQuantile Method}
\end{algorithm}

The {\em RandQuantile} method includes an option to {\em Extend} this set of intervals by creating one more interval on each side of the range.  This is done to allow for the possibility that there are test points beyond the range of values seen in training, and to avoid giving them zero probability. If {\em Extend} is specified as true, we specify three parameters $(q_{\min},q_{\max}, t)$ to determine the width of an additional bin on each side of the $[y_{\min}, y_{\max}]$ range defined by the training data.  This is a generalization of using the interquartile range (IQR) as a standardizing measure for the spread of a dataset.  We define a distance between the $q_{\min}$-th and $q_{\max}$-th quantiles and multiply it by a factor $t$.  So, setting $q_{\min} = .25,q_{\max}=.75, \text{and }t=1$ would be equivalent to adding an additional bin with width the size of the IQR to each side of the previous set of bins.

\begin{figure*}[t]
    \includegraphics[width=.24\textwidth]{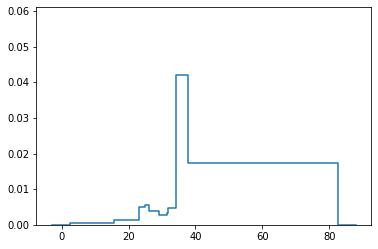}\hfill
    \includegraphics[width=.24\textwidth]{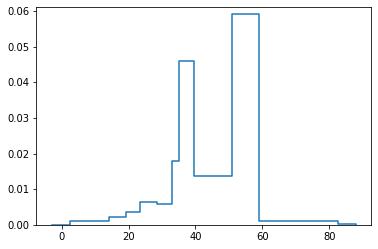}\hfill
    \includegraphics[width=.24\textwidth]{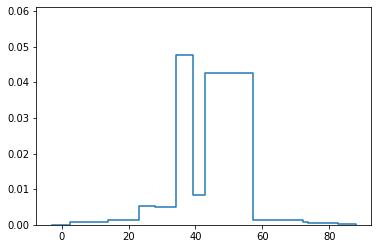}\hfill
    \includegraphics[width=.24\textwidth]{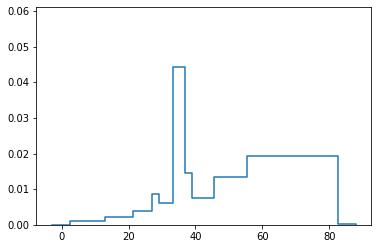}
    \caption{The densities predicted by 4 base classifiers on a single data point.}
    \label{fig:ind-densities}
    \includegraphics[width=.24\textwidth]{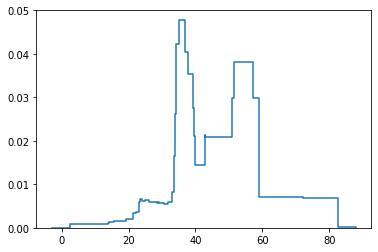}\hfill
    \includegraphics[width=.24\textwidth]{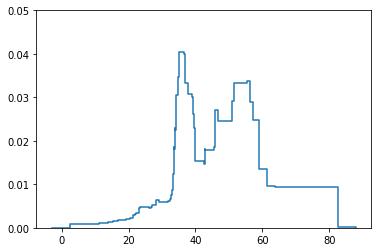}\hfill
    \includegraphics[width=.24\textwidth]{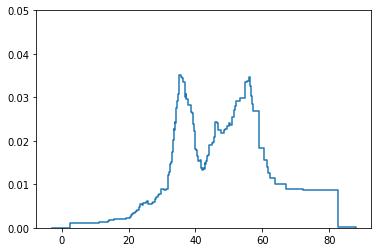}\hfill
    \includegraphics[width=.24\textwidth]{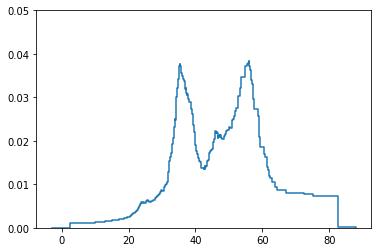}
    \caption{The averages of 5, 10, 25, and 50 base classifiers on the same data point}
    \label{fig:agg-densities}
    \end{figure*}

We illustrate the algorithm visually in Figure~\ref{fig:ind-densities} using the {\em Concrete} data set from the UCI repository.  In the top row we see the predicted densities from 4 different classifiers on the same data point.  These are the predictions from individual coarse classifiers on a particular test data point.  You can see that each classifier uses different intervals, and thus the densities look quite different from one another.  In the bottom row we see the predicted density on the same test data point when averaging $c$ coarse classifiers for $c=5,10,25, 50$.  As we combine multiple coarse densities, the resulting average density gets smoother.

\subsection{Structured Cross-Entropy}
\label{sec:strce}
One issue with applying a typical multi-classification algorithm to the intervals, is that, in general, multi-classification algorithms do not consider the {\em structure} of the classes.  Suppose we divide our target range into 10 intervals, numbered 1 through 10 from left to right.  If our algorithm puts very high probability on the outcome 3 when the right answer was 2, we are penalized just as much as if we had put equally high probability on 9. Loosely speaking, the standard cross-entropy loss only considers the probability placed on the {\em exact} right answer and does not give ``partial credit" for the probability placed {\em near} the correct answer.  The 10 different classes are equally different from one another, and no wrong answer is better than any other.

The {\em structured cross-entropy} loss function (\cite{lucena2022loss}) tries to mitigate this phenomenon by specifying a set of partitions of the classes.  Each partition represents a ``coarsening" of the state space, and is given a weight, such that the weights sum to one.  The (standard) cross-entropy is then calculated with respect to each of these partitions - i.e., the prediction is considered ``right" if the true answer and the prediction are in the same ``block" of the partition, and the probability of the block is the sum of the probabilities of the individual classes in the block.

\subsubsection{Example}
To make this more clear, we will walk through an example.  Suppose the target space is $\{1, 2, \ldots, 10\}$ and we expect an {\em ordinal} structure.  We could model this by specifying the partitions:
\begin{eqnarray*}
\mathcal{P}_0 &=&\{\{1\},\{2\},\{3\},\{4\},\{5\},\{6\},\{7\},\{8\},\{9\},\{10\}\} \\
\mathcal{P}_1 &=& \{\{1,2\},\{3,4\},\{5,6\},\{7,8\},\{9,10\}\} \\
\mathcal{P}_2 &= &\{\{1\},\{2,3\},\{4,5\},\{6,7\},\{8,9\},\{10\}\}  
\end{eqnarray*}
with corresponding weights $w_0,w_1, w_2>0$ such that $w_0+w_1+w_2=1$.

This structure means, in effect, that a distribution is credited, to some degree, for the probability placed on values ``adjacent" to the true value.  Suppose for a particular test data point, we predicted $p_a = (.1, .2, .4, .1, .05, .05, .025, .025, .025, .025)$ as our probability distribution over the 10 classes, and the true class was 2.  Since, the distribution $p_a$ assigns probability 0.2 to the outcome 2, the (standard) cross-entropy loss would be $-\log(.2) = 1.609$, as it only considers the probability placed on the exact correct class 2.

By contrast, the structured cross-entropy (with the structure defined above) averages the standard cross-entropy values taken with respect to the 3 partitions. $\mathcal{P}_0$ is the {\em singleton partition} and behaves just like the standard cross-entropy, with a result of $-\log(.2)$. For $\mathcal{P}_1$ the correct block was $\{1,2\}$. The distribution $p_a$ gave probability 0.1 to the outcome 1 and probability 0.2 to the outcome 2, so the probabillity of the block $\{1,2\}$ under $p_a$ is $.1+.2=.3$.  Thus the cross-entropy of the data point with respect to $p_a$ is  $-\log(.3)$.  For $\mathcal{P}_2$ the correct block was $\{2,3\}$ yielding a value of $-\log(.2+.4)$.   Averaging the three values together, using the corresponding weights, would yield $- w_0\log(.2)-w_1 \log(.3) - w_2 \log(.6) $.  For $w_0, w_1, w_2 = (.2, .4, .4)$ we would get a value of $1.008$.  Note that the structured cross-entropy considers the probability placed on values 1 and 3, even though the correct answer was 2.

Now consider another model, which on the same test data point predicts $p_b = (.1,.2,.1,.1,.1,.1,.1,.1,.05,.05)$.  Intuitively, this is a worse prediction than $p_a$, as it put less probability mass near the true answer of 2.  But from the point of view of standard cross-entropy, it is equally good: $-\log(.2) = 1.609$.  However, the structured cross-entropy (with the same partitions and weights) would be $- (.2)\log(.2) -(.4) \log(.1+.2) - (.4) \log(.2+.1)  = 1.285$.  The structured cross-entropy favors $p_a$ over $p_b$ whereas standard cross-entropy treats them as equivalent.

In the experimental section of this paper, we use the PRESTO algorithm with gradient boosted forests as the base classifier.  We consider variants trained under both the standard and structured cross-entropy loss functions.  We will see, both quantitatively and qualitatively, that the structured cross-entropy yields better results.

\subsubsection{Random Partitions}
In general, a {\em weighted partition set} is a set of partitions $\{\mathcal{P}_i\}$ of $\{1, \ldots, k\}$ with a corresponding set of weights $w_i$ which sum to one.  Since the weights sum to one, they could also be perceived as probabilities, and the entire object seen as a random variable taking values in the set of partitions.  Thus, it is also referred to as a {\em random partition}.  

\begin{table*}[t]
\begin{center}
\begin{tabular}{llllll}
\textbf{Dataset}  &\textbf{N} & \textbf{PRESTO} & \textbf{NGBoost} & \textbf{Best Prev Score} & \textbf{Best Prev Method}\\
\hline \\
Boston & 506 & \bf{2.564 $\pm$ 0.075} & \bf{2.43 $\pm$ 0.15} &\bf{2.37 $\pm$ 0.24} & \bf{Gaussian Process} \\
Concrete & 1030 & \bf{2.996 $\pm$ 0.034} & \bf{3.04 $\pm$ 0.09} & \bf{3.03 $\pm$ 0.11} & Gaussian Process \\
Energy & 768 & 0.922 $\pm$ 0.027 & \bf{0.60 $\pm$ 0.45} & \bf{0.60 $\pm$ 0.45} & \bf{NGBoost} \\
Kin8nm & 8192 &-0.550 $\pm$ 0.007 &-0.49 $\pm$ 0.02 & \bf{-1.20 $\pm$ 0.02} & \bf{Deep Ensembles} \\
Naval & 11934 & -5.228 $\pm$ 0.004 &-5.34 $\pm$ 0.04 & \bf{-5.87 $\pm$ 0.05} & \bf{Concrete Dropout}\\
Power & 9568 & \bf{2.525 $\pm$ 0.007} &2.79 $\pm$ 0.11 & 2.68 $\pm$ 0.05 & DistForest \\
Protein & 45730 & \bf{1.441 $\pm$ 0.006} &2.81 $\pm$ 0.03 & 2.59 $\pm$ 0.04 & DistForest \\
Wine & 1588 & \bf{0.781 $\pm$ 0.016} &0.91 $\pm$ 0.06 & 0.91 $\pm$ 0.06 & NGBoost \\
Yacht & 308 & 0.854 $\pm$ 0.088 & \bf{0.20 $\pm$ 0.26} & \bf{0.10 $\pm$ 0.26} & \bf{Gaussian Process} \\
YearMSD & 515345 & \bf{3.125$\pm$ NA} &3.43 $\pm$ NA & 3.35 $\pm$ NA & Deep Ensembles \\
\end{tabular}
\end{center}
\caption{NLL values on UCI Datasets.  The lowest value is in bold, as are those with standard errors that overlap the best value.} \label{nll-table}
\end{table*}

\begin{table}[t]
\begin{center}
\begin{tabular}{lllllll}
\textbf{Dataset}   & \textbf{PRESTO}  & \textbf{NGBoost} & \textbf{Prev Best}\footnotemark{} \\
\hline \\
Boston  & 3.522 $\pm$ 0.243  & 2.94 $\pm$ 0.53 &\bf{2.46 $\pm$ 0.32} \\
Concrete  & 5.524 $\pm$ 0.151  & 5.06 $\pm$ 0.61 &\bf{4.46 $\pm$ 0.29}  \\
Energy & 0.990 $\pm$ 0.033  & 0.46 $\pm$ 0.06 & \bf{0.39 $\pm$ 0.02}  \\
Kin8nm  & \bf{0.134 $\pm$ 0.001}  &0.16 $\pm$ 0.00 & 0.14 $\pm$ 0.00  \\
Naval  & 0.0013 $\pm$ 2e-5  & 0.00 $\pm$ 0.00 & 0.00 $\pm$ 0.00  \\
Power  & 3.378 $\pm$ 0.048  &3.79 $\pm$ 0.18 & \bf{3.01 $\pm$ 0.10}  \\
Protein  & \bf{3.371 $\pm$ 0.009}  &4.33 $\pm$ 0.03 & 3.60 $\pm$ 0.00  \\
Wine  & 0.579 $\pm$ 0.010  &0.63 $\pm$ 0.04 & \bf{0.50 $\pm$ 0.01}  \\
Yacht  & 2.399 $\pm$ 0.138  &0.50 $\pm$ 0.20 & \bf{0.42 $\pm$ 0.09}  \\
YearMSD  & 8.862 $\pm$NA  &8.94 $\pm$ NA & \bf{8.73 $\pm$ NA }
\end{tabular}
\end{center}
\caption{RMSE values on UCI Datasets.} \label{rmse-table}
\end{table}
There are many possible ways to design random partitions which capture ordinal structure.  We define here the {\em standard random ordinal partition} on $k$ classes with {\em singleton weight} $w_0$ and block size $s<(k/2)$ as follows:
\begingroup
\setlength{\medmuskip}{0mu}
\setlength{\thickmuskip}{0mu}
\setlength{\thinmuskip}{0mu}
\begin{eqnarray*}
\mathcal{P}_0 &=& \{\{1\},\{2\},\{3\}, \ldots, \{k\}\} \\
\mathcal{P}_1 &=& \{\{1, \ldots,s\},\{s+1,\ldots, 2s\}, \ldots \{\left\lfloor (k / s) \right\rfloor s+1,\ldots,k\}\} \\
\mathcal{P}_2 &=& \{\{1\}, \{2,\ldots,s+1\}, \ldots \{\left\lfloor (k / s) \right\rfloor s+2,\ldots,k\}\}\\
\mathcal{P}_3 &=& \{\{1,2\}, \{3,\ldots,s+2\}, \ldots \{\left\lfloor (k / s) \right\rfloor s+3,\ldots,k\}\}\\
\ldots \\
\mathcal{P}_s &=& \{\{1,2,\ldots,s-1\},\{s,\ldots,2s-1\}, \ldots \{\left\lfloor (k / s) \right\rfloor s, \ldots, k\}\} \\
w_i &=& (1-w_0)/s \text{\hspace{1cm} for $i = 1, \ldots, s$}
\end{eqnarray*}
\endgroup

\subsection{Usability}
PRESTO is available for public use via in the (pip-installable) StructureBoost package in Python.  It uses the standard \texttt{fit}/\texttt{predict} paradigm familiar to users of \texttt{scikit-learn}.  As such, it is simple to implement and accessible to non-experts.

\subsection{Computational Complexity}
This computation time for PRESTO is dominated by the training of the $m$ base classifiers, so it is equivalent to training $m$ multi-class gradient boosted forests (where the number of classes is approximately $r$ in the {\em RandQuantile} method).   Using the structured cross-entropy loss function slightly increases the training time, as the gradients are somewhat more complicated to compute.  However, this difference is typically not significant.  We trained a single forest with 200 trees, a max-depth of 7 and $r=25$ across 10 different splits of the {\em Power} dataset using both structured and stand took an average of 34.82 seconds while the equivalent forest trained with standard cross-entropy took an average of 34.22 seconds (on a 2019 MacBookPro).  Our experiments used 10 classifiers ($m=10$), so the computation time would be roughly 10 times that of a single forest.  Note also that although we did not implement any parallel computing techniques, the base classifiers can easily be trained in parallel, thereby reducing computation time considerably.

\section{EXPERIMENTS}
To evaluate PRESTO, we use a benchmark collection of 10 datasets from the UCI Machine Learning repository.  Overall, we rely heavily on the results presented in \cite{ngboost} and follow their methodology closely.  As such, we follow closely the protocol they used, which originated in ~\cite{pmlr-v37-hernandez-lobatoc15} and was used subsequently in \cite{gal-ghahramani}, \cite{concretedropout}, and \cite{lakshminarayanan-blundell}.  For each dataset, we set aside 10\% as a test set, and split the remainder 80-20 for training / validation.  The training/validation split is used for hyper-parameter optimization (including the number of boosting stages for each base classifier). Then the model is retrained on the combined training/validation data for the best choice of parameters and evaluated on the test set.   As in ~\cite{ngboost}, we ran 20 trials for each dataset, except for {\em Protein} (5) and {\em YearMSD} (1).  The latter was a particularly large dataset with a single train/test split, and therefore required special handling.

\subsection{Parameter Tuning}
The parameters of PRESTO fall into 3 main categories: 1) PRESTO-specific parameters including the interval selection method and number of classifiers, 2) parameters for the base classifier (gradient boosting) and 3) parameters regarding the structured entropy loss function.  As described below, we generally took an approach of simply choosing reasonable default values and did not do full parameter optimization.  Full details and code are available in the Supplementary Material

\subsubsection{PRESTO-Specific Parameters}
We used 10 classifiers for each dataset except for YearMSD, for which we used only a single classifier.  For the interval selection method, we used the {\em RandQuantile} method with $r=25$ and {\em Extend} with parameters $(.25, .75, .25)$ (i.e. extending 25\% of the InterQuartile Range).  There were 3 exceptions where the range and resolution of target values were known a priori: {\em Wine}, {\em Naval}, and {\em YearMSD}.  We used the {\em fixed-rss} method for {\em Naval} and the {\em fixed} method for {\em Wine}  and {\em YearMSD} .

\footnotetext{The previous best score was achieved by Gradient Boosting for all datasets except Protein and Wine, which were achieved by Random Forests.  The results for {\em Naval} were reported only two decimal places, thus the best score cannot be evaluated.}

\subsubsection{Gradient Boosting Parameters}
We chose learning rate values in relation to the size of the dataset.  We used a learning rate of 0.01 for the datasets with less than 2,000 datapoints, 0.05 for those between 2,000 and 12,000, .07 for {\em Protein} and .1 for {\em YearMSD}.  The {\em max-depth} was tuned on a single train/valid trial set, with the best value used for all subsequent trials.  The number of trees was tuned separately for each base classifier (forest) on each train/valid split.  The {\em YearMSD} contained a large number of rows (500K) and features (90) which precluded hyper-parameter optimization, so we arbitrarily chose a {\em max-depth} of 4 and randomly sampled .1 of the columns at each node for efficiency considerations.

\subsubsection{Structured Entropy Loss Parameters}
We trained PRESTO under two conditions: first, using a structured cross-entropy loss function (referred to as the {\em structured} variant), and second, using the standard cross-entropy loss function (referred to as the {\em standard} variant).  When using the structured variant, we used the standard ordinal random partition defined in Section~\ref{sec:strce} with $w_0 = .1$ and $s$ as the (rounded) square root of number of intervals.

\subsection{Negative Log Likelihood}
We use the average negative log-likelihood (a.k.a log-loss) measured on the test set as our primary metric of performance.  In Table~\ref{nll-table} we summarize the results of NGBoost and the suite of competitors that were evaluated in ~\cite{ngboost} and then add in the results for PRESTO (structured variant).  For brevity, we report the values only for NGBoost and the method that gave the best value for that particular dataset.  We will briefly summarize the competing methods here - for full details see ~\cite{ngboost} and the associated papers.  The {\em Concrete Dropout} method (\cite{concretedropout}) fits a neural network and uses a continuous relaxation to tune the dropout probability.  {\em Deep Ensembles} (\cite{lakshminarayanan-blundell}) fit multiple neural networks and then approximate a Gaussian mixture to obtain a probability distribution. The {\em Gaussian Processes} method (\cite{gpbook}) used a relevance detection kernel optimized by gradient descent and employing a variational inference technique from \cite{titsias}.  The {\em DistForest} method (\cite{schlosser}) used forests of 200 trees to estimate the parameters of a Normal distribution.

We see that PRESTO performs competitively overall, scoring best on 5 out of the 10 datasets and statistically close on a 6th.  It is particularly effective on the larger datasets:  the improvement on the Protein dataset is particularly notable in its magnitude.  It outperforms NGBoost on 6 out of the 10 datasets - with NGBoost typically achieving better results on the smaller datasets.  This is not surprising - PRESTO should have lower bias and higher variance and therefore require more data.

\begin{table}[t]
\begin{center}
\begin{tabular}{llllll}
\textbf{Dataset}  & \textbf{20\%} & \textbf{50\%} & \textbf{80\%} & \textbf{90\%} & \textbf{95\%}\\
\hline \\
Boston  & 0.249 & 0.632 & 0.934 & 0.986 & 0.989 \\
Concrete  & 0.293 & 0.689 & 0.953 & 0.987 & 0.992 \\
Energy & 0.392 & 0.851 & 0.985 & 0.99 & 0.995 \\
Kin8nm  &0.246 & 0.621 & 0.913 & 0.974 & 0.993 \\
Naval & 0.544 & 0.883 & 0.958 & 0.961 & 0.962  \\
Power  & 0.261 & 0.621 & 0.898 & 0.961 & 0.984 \\
Protein & 0.283 & 0.624 & 0.881 & 0.949 & 0.977 \\
Wine  & 0.498 & 0.740 & 0.874 & 0.928 & 0.956 \\
Yacht  & 0.355 & 0.721 & 0.984 & 0.995 & 0.995 \\
YearMSD & 0.197 & 0.500 & 0.800 & 0.902 & 0.948  
\end{tabular}
\end{center}
\caption{Coverage Rate on UCI Datasets} \label{pi-table}
\end{table}

\subsection{Point Estimation}
While PRESTO is intended to yield a full conditional density rather than a single point estimate, it is still possible to provide a point estimate by calculating the mean of the predicted density.  In Table~\ref{rmse-table} we report performance results for predicting point estimates as measured by RMSE.  Again, we summarize the results provided in \cite{ngboost} and add in the results for PRESTO.  In general, the methods designed to predict point estimates outperform both PRESTO and NGBoost as measured by RMSE.  Nevertheless, PRESTO gets the best results on 2 out of the 9 datasets.  In head-to-head with NGBoost, PRESTO gets better results on the 5 datasets containing more than 1500 datapoints, while NGBoost outperforms on the 4 smaller datasets.  One possible explanation for the lack of performance on point prediction regards the nature of density prediction.  Since there is a high penalty for putting a low probability on an observed value, density predictions must be more conservative and put {\em some} probability mass on the most extreme values.  Consequently, these ``fat tails" influence the mean, which hurts performance on the RMSE metric.

\subsection{Prediction Intervals}
One valuable aspect of predicting a full conditional density is the ability to specify a range of values and a probability that the target will fall within that range.  Of course, such intervals are only useful insofar as the coverage guarantees prove to be true in practice.  To test, we created $(1-\alpha)$-prediction intervals by computing the $\alpha/2$ quantile and the $1-(\alpha/2)$ quantiles of the predicted (conditional) density and evaluating what percentage of time the true target value fell into the associated interval.  This was done for $\alpha =0.8, 0.5, 0.2, 0.1, 0.05$ to create prediction intervals at coverages of (20\%, 50\%, 80\%, 90\% and 95\%) respectively.  The results are in Table~\ref{pi-table}.  For all but the {\em YearMSD} dataset, we see that the 95\% prediction intervals contained the true value more than 95\% of the time, and that this pattern applied to the other prediction intervals as well.  In fact, the data suggest that the intervals are wider than necessary, and perhaps too conservative.  This may be another consequence of minimizing the negative log-likelihood, which encourages ``fatter tails" due to the extreme penalties of low densities.  Nevertheless, this analysis demonstrates that, in practice, the prediction intervals give at least the promised coverage, which is a valuable property.  For the {\em YearMSD} dataset, we see that the empirical coverage tracks very closely to the purported coverage.  This may be a consequence of the large size of that dataset.
\begin{table}[t]
\begin{center}
\begin{tabular}{llll}
\textbf{Dataset}  &\textbf{N} & \textbf{PRESTO-struc} & \textbf{PRESTO-std} \\
\hline \\
Boston & 506 & 2.564 $\pm$ 0.075 & 2.635 $\pm$ 0.076 \\
Concrete & 1030 & 2.996 $\pm$ 0.034 &  3.054 $\pm$ 0.035 \\
Energy & 768 & 0.922 $\pm$ 0.027 &  0.953 $\pm$ 0.027  \\
Kin8nm & 8192 &-0.550 $\pm$ 0.007 &-0.500 $\pm$ 0.006 \\
Naval & 11934 & -5.228 $\pm$ 0.004 & -5.199 $\pm$ 0.004  \\
Power & 9568 & 2.525 $\pm$ 0.007 & 2.546 $\pm$ 0.007 \\
Protein & 45730 & 1.441 $\pm$ 0.006 & 1.447$\pm$ 0.006 \\
Wine & 1588 & 0.781 $\pm$ 0.016 &0.784 $\pm$ 0.015  \\
Yacht & 308 & 0.854 $\pm$ 0.088 & 0.933 $\pm$ 0.089  \\
YearMSD & 515345 & 3.125 $\pm$ NA &3.167 $\pm$ NA  
\end{tabular}
\caption{NLL values for PRESTO Variants} \label{table-strent}
\end{center}
\end{table}

\subsection{Loss Function Evaluation}
\begin{figure*}[t]
    \includegraphics[width=.24\textwidth]{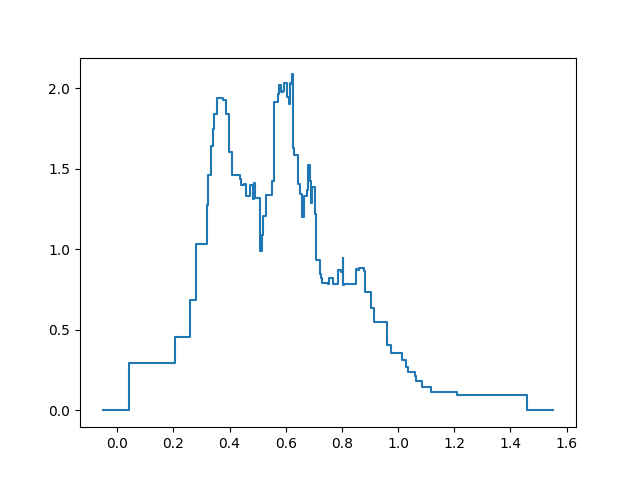}\hfill
    \includegraphics[width=.24\textwidth]{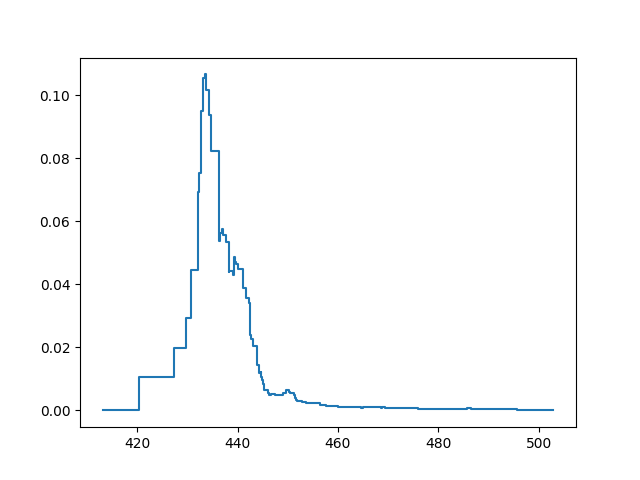}\hfill
    \includegraphics[width=.24\textwidth]{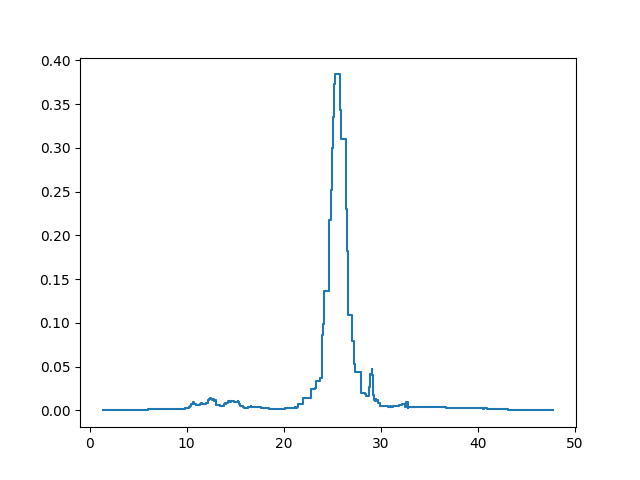}\hfill
    \includegraphics[width=.24\textwidth]{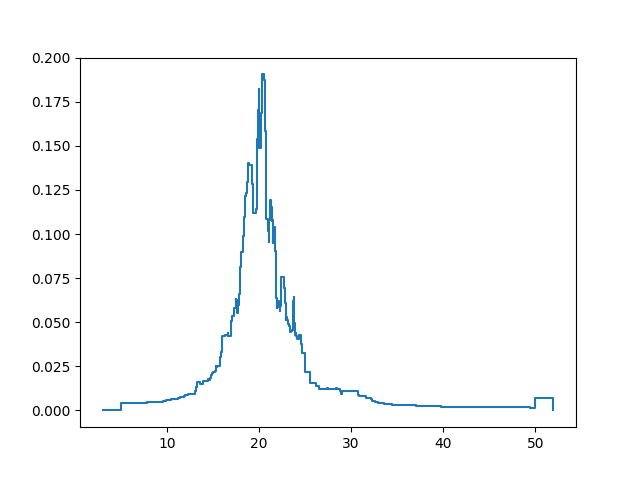}
    \includegraphics[width=.24\textwidth]{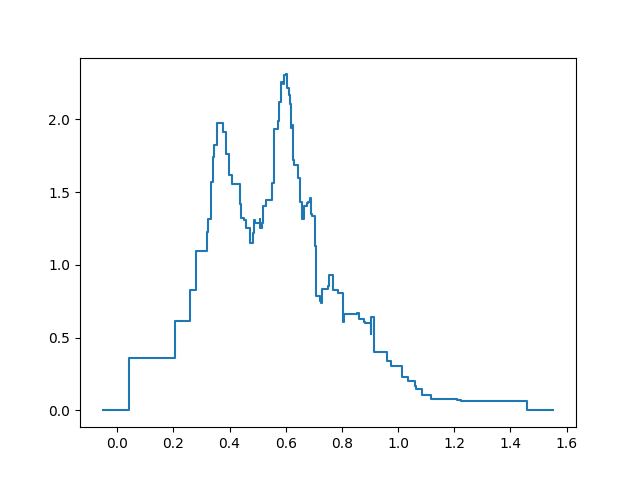}\hfill
    \includegraphics[width=.24\textwidth]{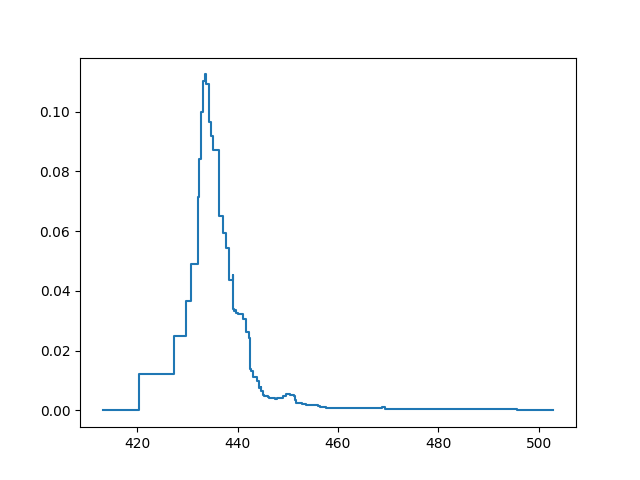}\hfill
    \includegraphics[width=.24\textwidth]{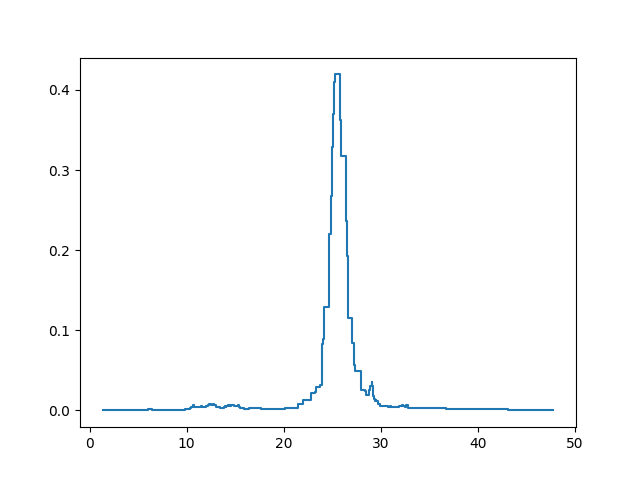}\hfill
    \includegraphics[width=.24\textwidth]{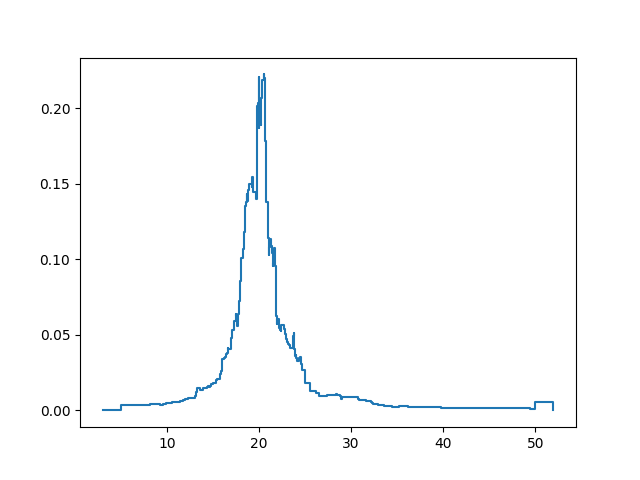}
    \caption{Top row: Predicted densities using standard variant from {\em kin8nm}, {\em Power}, {\em Energy}, and {\em Boston}  (left to right).
    Bottow row: same densities using structured variant.}
    \label{fig:densities1}
\end{figure*}

To assess the impact of using the structured entropy loss function, we compare the structured and standard variants of PRESTO.  The results are in Table~\ref{table-strent}.  The structured variant consistently, if modestly, outperforms the variant trained with the standard cross-entropy across all 10 datasets.  Therefore, a simple one-sided sign test against the null hypothesis of no difference indicates a p-value of $(1/2^{10}) <.001$.  On a more qualitative level, the densities resulting from the structured variant are smoother.  This is demonstrated in Figure~\ref{fig:densities1}. The top row shows predicted densities from models using the standard variant (from the {\em Kin8nm}, {\em Power}, {\em Energy} and {\em Boston} datasets, from left to right).  The bottom row shows the corresponding predictions from the structured model.  Visually, we can see that the densities on the bottom row are smoother and less jagged.

\begin{figure}[b]
    
    \includegraphics[width=.24\textwidth]{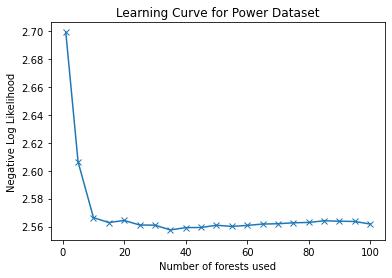}\hfill
    \includegraphics[width=.24\textwidth]{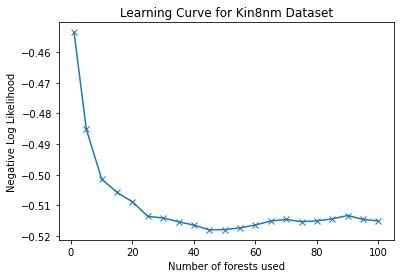}
     \caption{Learning Curves for {\em Power} and {\em Kin8nm}.}
     \label{fig:lc}

\end{figure}

\subsection{Learning Curves}
In our experiments, we (somewhat arbitrarily) chose 10 as the default number of classifiers (forests).  Since we average the classifiers in our ensemble, we would expect that performance should improve and then level off as we add more classifiers, but that we should not overfit by adding more classifiers.  (The individual classifiers, of course, must be trained carefully to avoid overfitting, but that is a different issue.) To explore this empirically we ran PRESTO with 100 forests on a single trial for the {\em Power} and {\em Kin8nm} datasets and plotted the the performance (negative log-likelihood) as a function of the number of forests used.  The results are in Figure~\ref{fig:lc}.  For the Power dataset, we see that performance quickly reaches its optimum level: between 10 and 100 forests the performance level is essentially constant.  For the Kin8nm dataset, it takes longer to level off: performance is notably better at 25 forests than at 10.  This suggests we may be able to improve performance on some datasets by increasing the number of forests.  The results from both datasets suggest that overfitting due to a high number of forests is not a concern.

\section{SUMMARY}
PRESTO is shown to be an effective method for probabilistic regression.  It makes minimal assumptions about the form of the predicted densities, effectively learning the shape from the data.  It performs competitively on a benchmark set of datasets against a suite of competing algorithms, yielding the best result on 5 out of the 10 datasets.  It outperforms considerably on the largest datasets (of sizes 45K and 515K), while its weakest results came on smaller datasets ($<$1K data points).  Associated prediction intervals are empirically shown to have appropriate coverage.  The intervals were too conservative on most of the datasets, providing a higher rate of coverage than expected, and suggesting that narrower intervals could have provided the promised coverage.  However, for the largest dataset (515K), the intervals were quite well calibrated, suggesting that the precision may improve given enough data - but more research is needed to validate this.  Finally, it can be useful as a tool for inference, as it learns the form and shape of the conditional density, potentially yielding valuable insights.  It employs the standard fit/predict paradigm and is therefore easy to use by experts and non-experts alike.

There are several potential avenues for future work in this area.  For example, it would be interesting to explore why the prediction intervals are wider than necessary.  Potentially this may arise from a lack of calibration in the base classifiers.  If that is the case, perhaps performing post-hoc calibration on the individual base classifiers may improve the quality of those intervals (as well as the general performance).  Moreover, since multimodal predictions are not uncommon, there is also the potential to provide disconnected prediction regions - that is, choosing the intervals of highest density rather than just using the center of the distribution.  Such prediction regions would be narrower (in aggregate) than the central quantile approach used in this paper.

\bibliography{probreg_arxiv}

\end{document}